\pdfoutput=1

\documentclass{article}
\usepackage[utf8]{inputenc}

\title{Synthesizing Safe Policies under Probabilistic Constraints with Reinforcement Learning and Bayesian Model Checking}
\author{Lenz Belzner and Martin Wirsing}
\date{}

\usepackage{natbib}
\usepackage{graphicx}
\usepackage{todonotes}
\presetkeys%
    {todonotes}%
    {inline}{}
\usepackage{amssymb}
\usepackage{algorithm}
\usepackage[noend]{algpseudocode}
\usepackage{algorithmicx}
\usepackage{amsmath}
\usepackage{dirtytalk}

\usepackage{caption}
\usepackage{subcaption}
\usepackage{hyperref}

\usepackage{comment}

\setcitestyle{numbers}
\setcitestyle{comma}
\setcitestyle{square}

\newtheorem{fact}{Fact}

\begin{document}

\maketitle

\noindent {\footnotesize \textbf{Draft.} Final Version: Lenz Belzner and Martin Wirsing. Synthesizing Safe Policies under Probabilistic Constraints with Reinforcement Learning and Bayesian Model Checking. Science of Computer Programming, 2021, 102620, ISSN 0167-6423.}
\-\\

\begin{abstract}
    We propose to leverage epistemic uncertainty about constraint satisfaction of a reinforcement learner in safety critical domains. We introduce a framework for specification of requirements for reinforcement learners in constrained settings, including confidence about results. We show that an agent's confidence in constraint satisfaction provides a useful signal for balancing optimization and safety in the learning process.
\end{abstract}

\section{Introduction}

Reinforcement learning enables agents to automatically learn policies maximizing a given reward signal. Recent developments combining reinforcement learning with deep learning have had great success in tackling more and more complex domains, such as learning to play video games based on visual input or enabling automated real-time scheduling in production systems \cite{arulkumaran2017deep,waschneck2018optimization}.

Reinforcement learning also provides valuable solutions for systems operating in non-deterministic and partially known environments, such as autonomous systems, socio-technical systems and collective adaptive systems (see e.g. \cite{Onplan15,HoelzlG16,CardosoRHKP18,MahfoudhSBA18}). 
However, it is often difficult to ensure the quality and the correctness of reinforcement learning solutions \cite{russell2015research,amodei2016concrete}. 
In many applications, learning is focusing on
achieving and optimizing system behavior but not on guaranteeing the safety of the system (see e.g. also~\cite{Onplan15,HoelzlG16,CardosoRHKP18}).

Optimizing for both functional effectiveness and system safety at the same time poses a fundamental challenge: If maximizing return and constraint satisfaction are interfering, what should the learner optimize in a given situation? That is, besides the fundamental dilemma of exploration and exploitation, the learner now faces an additional choice to be made: When to optimize return, and when to optimize feasibility in case it is not possible to optimize both at the same time?

This fundamental question has been addressed in many related works, which can be categorized in three broad classes: Learning safe behavior using a given or learned model of the environment~\cite{junges2016safety, haesaert2018temporal, hasanbeig2020cautious}, learning shields in addition to a policy assuring that only safe actions are executed~\cite{alshiekh2017safe, bharadwaj2019synthesis, avni2019run, jansen_et_al:LIPIcs:2020:12815} or reward-shaping methods that try to balance optimization of return and costs incurred by constraint violation~\cite{raybenchmarking, chow2017risk, chow2018lyapunov, fan2019safety}.

Typically satisfaction probability is estimated via maximum likelihood, ignoring the learner's uncertainty about intermediate estimates in the empirical learning process. We argue that when using reinforcement learning for policy synthesis, additionally requiring a specification of \textit{confidence} in the result is of high importance as reinforcement learning is an empirical technique relying on finite data points. We also argue that using a learner's confidence in requirement satisfaction provides a highly informative signal for effectively exploring the Pareto front of return optimization and safety.

To this end, we propose \textit{Policy Synthesis under probabilistic Constraints} (PSyCo), a systematic method for specifying and implementing agents that shape rewards dynamically over the learning process based on their confidence in requirement satisfaction. The basic idea is to emphasize return optimization when the learner is confident, and to focus on satisfying given constraints otherwise. This enables to explicitly distinguish requirements wrt. aleatoric uncertainty that is inherent to the domain, and epistemic uncertainty arising from an agent's learning process based on limited observations.

For implementing PSyCo's abstract design we propose \textit{Safe Neural Evolutionary Strategies} (SNES) for model-free learning of safe policies wrt. given finite-horizon specifications. SNES leverages Bayesian model checking while learning to adjust the Lagrangian of a constrained optimization problem according to the learner's current confidence in specification satisfaction. The model-free formulation allows to synthesize safe policies in domains where only a generative model of the environment is available, enabling learning when the model is unknown to the agent or where analytical computation of solutions wrt. transition dynamics is intractable for closed form representations. Also, the model-free formulation enables fast inference (i.e. action selection) at runtime, rendering our approach feasible for real-time application domains.




The paper makes the following contributions.
\begin{itemize}
    \item \textit{Policy Synthesis under probabilistic Constraints} (PSyCo), a systematic method for shaping rewards dynamically over the learning process based on the learner's confidence in requirement satisfaction. PSyCo accounts for empirical policy synthesis and verification based on finite observations by including requirements on confidence in synthesized results.
    \item \textit{Safe Neural Evolutionary Strategies} (SNES) for learning safe policies under probabilistic constraints. SNES leverages online Bayesian model checking to obtain estimates of constraint satisfaction probability and a confidence in this estimate.
    \item We empirically evaluate SNES showing it is able to synthesize policies that satisfy probabilistic constraints with a required confidence. Code for our experiments is available at \url{https://github.com/lenzbelzner/psyco}.
\end{itemize}

The paper is structured as follows: In Section 2 we introduce the PSyCo method. Section 3 describes safe policy synthesis with the Safe Neural Evolutionary Strategy SNES. In Section 4 we present the results of  two experiments, the so-called Particle Dance and Obstacle Run case studies. Sections 5 and 6 discuss related work and the limitations of our approach. Finally, Section 7 gives a short summary of PSyCo and addresses further work.  

\section{The PSyCo Method for Safe Policy Synthesis under Probabilistic Constraints}

Our approach to safe policy synthesis comprises three phases: System specification as constrained Markov decision process with goal-oriented requirements, system design and implementation via safe policy synthesis, and verification by Bayesian model checking. As we will see, we use Bayesian model checking in two ways: To guide the learning process towards feasible solutions, and to verify synthesized policies.

\subsection{PSyCo Overview}

PSyCo comprises the four components.
\begin{itemize}
    \item A domain specification given by a Markov decision process. We emphasize that our approach is model-free, therefore only requiring a generative model of the transition dynamics. The model-free formulation allows to synthesize safe policies when the dynamics model is unknown or where analytical computation of solutions wrt. transition dynamics is intractable for closed form representations.
	\item A set of goal-oriented requirements including optimization goals and probabilistic safety constraints. We restrict our further discussion to a single optimization goal and a single safety constraint for the sake of simplicity. We think that extending our results to sets of constraints is straightforward. We derive a cost function from requirements, effectively turning the domain MDP into a constrained Markov decision process (CMDP).
	\item A safe reinforcement learning algorithm $L$ yielding the parameters of a policy.
	\item A verification algorithm $V$ to check constraint satisfaction of the learned policy in the given CMDP.
\end{itemize}

PSyCo leverages a learning algorithm $L$ for synthesizing safe policies wrt. rewards, costs and constraints (i.e. optimizing goals and probabilistic constraints), and a verification algorithm $V$ for statistically verifying synthesized policies.

We optimize the parameters $\theta \in \Theta$ of the policy wrt. rewards and costs of the derived constrained CMDP $m \in M$ with the safe reinforcement learning algorithm $L$, taking the given probabilistic requirement $\varphi \in \Phi$ into account.
\begin{equation}
L : M \times \Phi \rightarrow \Theta
\end{equation}

We verify the optimized parameters of the synthesized policy wrt. the given CMDP and the constraint specification.
\begin{equation}
V : M \times \Phi \times \Theta \rightarrow \mathbb{B}
\end{equation}

Given a CMDP $m$ and a constraint $\varphi$, PSyCo works by learning and verifying a policy as follows (where $\theta = L(m, \varphi)$).

\begin{equation}
\mathrm{PSyCo} : M \times \Phi \rightarrow \mathbb{B}
\end{equation}
\begin{equation}
\mathrm{PSyCo}(m, \varphi) = V(m, \varphi, L(m, \varphi)) = V(m, \varphi, \theta)  
\end{equation}

\subsection{System Specification: Constrained Markov Decision Processes and Goal-Oriented Requirements}
\label{sec:specification}

\paragraph{Domain Specification as Constrained MDP}
The specification for a particular domain is given by a set $S$ of states, a distribution over initial states $\rho : p(S)$, a set $A$ of actions, and a reward function $R : S \times A \times S \rightarrow \mathbb{R}$ encoding optimization goals for the agent.

We assume the domain has probabilistic transition dynamics $T : p(S \vert S, A)$. Note that $T$ may be \textit{unknown} to the agent. 

A specification comprises a set of requirements or constraints given in a bounded variant of PCTL that we define below. We also define a way to transform these constraints into a cost function $C : S \times A \times S \rightarrow \mathbb{R}$. $(S, A, T, R, C, \rho)$ constitutes a constrained Markov decision process (CMDP) \cite{altman1999constrained}. 

An episode $\vec{e} \in E$ is a finite sequence of transitions $(s_i, a_i, s_{i + 1}, r_i, c_i)$, $s_i, s_{i + 1} \in S$, $a_i \in A, r_i = R(s_i, a, s_{i + 1}), c_i = C(s_i, a, s_{i + 1})$ in the CMDP. The sequence $\vec{s} = s_0, ..., s_{n - 1}$  denotes the  path $s(\vec{e})$ of that episode and $Path_n$ denotes the set of all paths of length $n$.

In a CMDP, the task is to synthesize a deterministic, memoryless policy $\pi : S \rightarrow A$ that maximizes reward while minimizing costs. We will formally specify our particular task after introducing our safety specifications and their transformation into cost functions below.

We emphasize that our approach is model-free, therefore only requiring a generative model of the transition dynamics $T$. The model-free formulation allows us to synthesize safe policies when the dynamics model is unknown or where analytical computation of solutions wrt. transition dynamics is intractable for closed form representations. Note that as we only rely on sampling the domain for policy synthesis in the following, even partial observations and a generative model of the reward function are sufficient for our approach to work.

Note that CMDPs are not restricted to a single cost function in general, however in this paper we restrict ourselves to a single cost function for sake of simplicity. We think that our results could be extended to sets of cost functions straightforwardly.

\paragraph{Requirements}  

We consider two kinds of goals: \textit{optimization goals} and \textit{safety constraints}. Optimization goals are soft constraints and maximize an objective function,
constraints are behavioral goals which impact the possible behaviors of the system, similar to e.g. maintain goals (which restrict the behavior of the system) and achieve goals (which generate behavior), see KAOS \cite{DardenneLF93}. 
In our setting we relate the optimization goal with the rewards of the MDP and require it to maximize the return: 
\begin{equation}
\label{eq:optimize}
\mathbf{Goal} \text{ Optimize } \mathit{ Return} : \max \mathbb{E}(\mathcal{R})
\end{equation}
where the return $\mathcal{R}$ is the cumulative sum of rewards $\mathcal{R} = \sum_{i = 0}^{|\vec{e}|-1} r_i$  in an episode $\vec{e}$ and $\mathbb{E}(\mathcal{R})$ denotes the expectation of the return. 

A suitable logic to express safety constraints in our setup is probabilistic computation tree logic (PCTL), allowing to specify constraints on satisfaction probabilities as well as bounding costs that may arise in system execution~\cite{pnueli1977temporal, baier2008principles}.
We interpret PCTL formulas as bounded by the length of an episode $n \in \mathbb{N}$ 
and consider typically formulas of the form 
\begin{equation}
\label{eq:formula}
\mathbb{P}_{\geq p_\mathrm{req}} \left( \square \phi \right)
\end{equation}
where $\square \phi$ is a so-called path formula and $\phi$ is a propositional state formula built according to the syntax for CTL state formulas ($a$ is an atomic state proposition, $\phi_1, \phi_2$ are state formulas).
\begin{equation}
    \phi = \mathrm{true} ~\vert~ a ~\vert~ \phi_1 \wedge \phi_2 ~\vert~\neg \phi
\end{equation} 

Formula \ref{eq:formula} states that with at least probability $p_\mathrm{req}$, the constraint $\phi$ holds in every state of the path of the episode (i.e. $\forall k, 0 \leq k \leq n : s_k \models \phi$). 
For the formal semantics and a more general treatment of other formulas in PCTL, we refer the reader to the appendix~\ref{sub:PCTL}.

Given that reinforcement learning is an empirical approach for synthesizing policies, it is reasonable to require a confidence in any synthesized result. Therefore we extend our constraints by an explicit operator 
$\mathbb{C}_{c_{req}}$, requiring the rate of false positive and false negative verification results to be bounded by $1-c_{req}$. This allows to distinguish between aleatoric (i.e. domain-inherent, irreducible) and epistemic (i.e. agent-inherent, reducible) uncertainty in the specification of requirements. PSyCo enables to specify requirements for aleatoric uncertainty by bounding the probability of a constraint violation and epistemic uncertainty by specifying a bound on confidence.
We denote our requirements wrt. constraint violation probability and agent confidence by
\begin{equation}
\mathbb{P}_{\geq p_\mathrm{req}} \left( \square \phi \right) \textrm{ and }  \mathbb{C}_{\geq c_\mathrm{req}}
\end{equation}
requiring that $\square \phi$ holds with probability of at least $p_{req}$ and confidence of at least $c_{req}$ ($p_{req}, c_{req} \in [0,1]$).

Let us consider how to derive a cost function $C$ from a given constraint. For any propositional state formula $\phi$ we 
define the cost function $C : S \times A \times S \rightarrow \mathbb{R}$ such that
\begin{equation}
C_\phi(s, a, s')
\begin{cases}
= 0               & \text{if } s' \vDash \phi  \\
> 0               & \text{otherwise}
\end{cases}
\end{equation}

Note that another value of $C$ for a violation of $\phi$ could  be given by a more general function of the post-state $s'$, giving the possibility to quantify the severeness of a present violation. 


The cumulative cost $\mathcal{C}$ of an episode $\vec{e}$ of length $n$ with path $\vec{s}$ for a safety constraint $\square \phi$ is defined by the sum of violations of $\phi$ over the course of the episode.
\begin{equation}
\label{eq:formula1}
    \mathcal{C}_{\square \phi, \vec{s}} = c_\phi(s_0) + \sum_{i = 0}^{n - 1} C_{\phi}(s_t, a, s_{t + 1})
\end{equation}
where the cost $c_\phi(s_0)$ of the initial state of an episode is computed with the function $c_\phi : S \rightarrow \mathbb{R}$ defined by
\begin{equation}
c_\phi(s) =
\begin{cases}
0               & \text{if } s \vDash \phi  \\
1               & \text{otherwise}
\end{cases}
\end{equation}

In this way, we get a measure of how often an violation actually happened. This provides a more fine-grained learning signal to the learning agent than a strictly binary reward for satisfaction or violation. 

Also note that the cumulative cost of an episode is zero exactly if no state in the episode violates $\phi$. Then the relation of a given safety constraint and the cumulative cost of an episode $\vec{e}$ with path $\vec{s}$ is given by
\begin{equation}
    \vec{e} \models \square \phi \text{ iff } \mathcal{C}_{\square \phi, \vec{s}} = 0
\end{equation}

For more general path formulas and details see appendix \ref{sub:PCTL} and \ref{sub:cost}.

\subsection{Abstract Design: Safe Reinforcement Learning Algorithm L}

The task defined by a PSyCo specification is to synthesize a deterministic, memoryless policy $\pi : S \rightarrow A$ that optimizes the following constrained optimization problem for bounded episodes.
\begin{equation}
\label{eq:target}
    \max \mathbb{E}(\mathcal{R}) \textrm{ s.t. } \mathbb{P}_{\geq p_\mathrm{req}}(\mathcal{C_{\varphi}} = 0) \textrm{ and }  \mathbb{C}_{\geq c_\mathrm{req}}
\end{equation}

We now discuss the learning algorithm L for synthesizing a policy optimizing its parameters wrt. this constrained optimizaton problem.

We denote executing a policy parameterized by $\theta \in \Theta$ in a CMDP $m \in M$ as follows, yielding a distribution over episodes.
\begin{equation}
	m : p( E \vert \Theta)
\end{equation}
We sample episodes from the distribution $m$, which we denote as follows.
\begin{equation}
	\vec{e} \sim m(\theta)
\end{equation}

Note that we overload $m$ to describe both the CMDP tuple and the probability distribution the CMDP yields when being executed with a policy and constraints. 

A policy that optimizes the constrained optimization problem (c.f. Equation \ref{eq:target}) can be synthesized with safe reinforcement learning \cite{garcia2015comprehensive}. One approach is to formulate the problem as a Lagrange function and use it as a reward function for a reinforcement learning algorithm \cite{altman1999constrained}. Given an episode sampled from an MDP, we can compute cumulative return $\mathcal{R}$ and cost $\mathcal{C}_\varphi$ for a given episode and a given path formula $\varphi$ as defined in Section \ref{sec:specification}. In general, we can transform the problem
\begin{equation}
    \max \mathcal{R} \textrm{ s.t. } \mathcal{C}_\varphi = 0
\end{equation}
to its Lagrange formulation
\begin{equation}
\label{eq:dual}
    \max \mathcal{R} - \lambda \mathcal{C}_\varphi
\end{equation}
where $\lambda \in \mathbb{R}^+$ is a Lagrangian multiplier \cite{beavis1990optimisation}. Without loss of generality, we use an alternative formulation of Eq. \ref{eq:dual} where $\lambda \in (0, 1)$.
\begin{equation}
    \max \mathcal{R} - (1 - \lambda) \mathcal{C}_\varphi
\end{equation}

We outline the general process of safe RL with function approximation in Algorithm \ref{alg:Safe RL}. Note that the expectation is not given in line 7 due to sampling of episodes, returns and costs. Also note that Algorithm \ref{alg:Safe RL} does not ensure safety while learning, but only when converging to a solution of the Lagrangian. The key challenge in this approach is to determine an appropriate $\lambda$ and to adjust it effectively over the learning process.
\begin{algorithm}[H]
	\begin{algorithmic}[1]
		\Procedure {Safe RL}{CMDP $m \in M$, path formula $\varphi$}
		\State initialize parameters $\theta$
		\While {learning}
		
		\State generate episodes by sampling from $m$
		\State determine return $\mathcal{R}$ and cumulative cost $\mathcal{C}_\varphi$
		\State determine $\lambda$
		\State update $\theta$ wrt. $\max_\theta \lambda \mathcal{R} - (1 - \lambda) \mathcal{C}_\varphi$
		
		\EndWhile
		\EndProcedure
	\end{algorithmic}
	\caption{Safe RL}
	\label{alg:Safe RL}
\end{algorithm}

\subsection{Verification: Bayesian Model Checking Algorithm V}

The empirical nature of reinforcement learning necessitates quantification of confidence about properties of learned policies. We resort to statistical model checking~\cite{legay2010statistical} to quantify confidence and verify policies accordingly, in particular to Bayesian model checking (BMC) \cite{jha2009bayesian, zuliani2010bayesian, belzner2017bayesian}. Bayesian model checking models epistemic uncertainty about satisfaction probability via sequential Bayesian update of the posterior distribution. We leverage BMC in two ways: To guide the learning process towards feasible solutions, and to verify synthesized policies.

Executing a policy in $m$ generating an episode either satisfies or violates a given bounded path formula without probabilistic operator. Thus, we can treat the generation of multiple episodes as Bernoulli experiment with a satisfaction probability $p_\mathrm{sat}$. We want to estimate this probability in order to check whether it complies with our probabilistic constraint $p_\mathrm{req}$, that is $p_\mathrm{sat} \geq_? p_\mathrm{req}$. 

Rather than doing a point estimate of $p_\mathrm{sat}$ via maximum likelihood estimation we assign a plausibility to each possible $p_\mathrm{sat} \in (0, 1)$, yielding a Bayesian estimate. We assign a prior distribution to all possible values of $p_\mathrm{sat}$, and then compute the posterior distribution based on the observations $O$ of cost constraint satisfaction or violation. The posterior distribution allows us to quantify our confidence in whether a given required satisfaction probability is met.

In general, the posterior is proportional to the prior $P(p)$ and the likelihood of the observations given this prior.
\begin{equation}
	\label{eq:bayes}
	P(p | O) \propto P(O | p) P(p)
\end{equation}

In the particular case of a Bernoulli variable, the conjugate prior is the Beta distribution \cite{diaconis1979conjugate}, meaning that prior and posterior distribution are of the same family (the Beta distribution in our case). The Beta distribution is defined by two parameters $\alpha, \beta \in \mathbb{N}^+$, which are given by the observed count of positive and negative results of the Bernoulli experiment. We use a uniform prior $\alpha, \beta = 1$ over possible values of $p_\mathrm{sat}$, assigning the same plausibility to all possible values before observing any data. This yields the following equality when assuming $s$ satisfactions and $v$ violations of a given constraint.
\begin{equation}
\label{eq:beta}
P(p_\mathrm{sat} | s, v) = \mathrm{Beta}(s + 1, v + 1)
\end{equation}

This update yields a posterior distribution over the possible values of $p_\mathrm{sat}$ given the observation of satisfaction and violation. We can now compute the probability mass $c_\mathrm{sat}$ of this posterior that lies above the required probability $p_\mathrm{req}$ to obtain a confidence about the current system satisfying our probabilistic constraint.
\begin{equation}
\label{eq:c_sat}
	c_\mathrm{sat} = \int\limits_{p_\mathrm{req}}^{1} P(p_\mathrm{sat}) d p_\mathrm{sat} = 1 - \mathrm{Beta}(s + 1, v + 1).\mathrm{cdf}(p_\mathrm{req})
\end{equation}
Here, $\mathrm{cdf}$ denotes the cumulative density function of the Beta distribution.

\begin{figure}
\centering
\includegraphics[width=.8\textwidth]{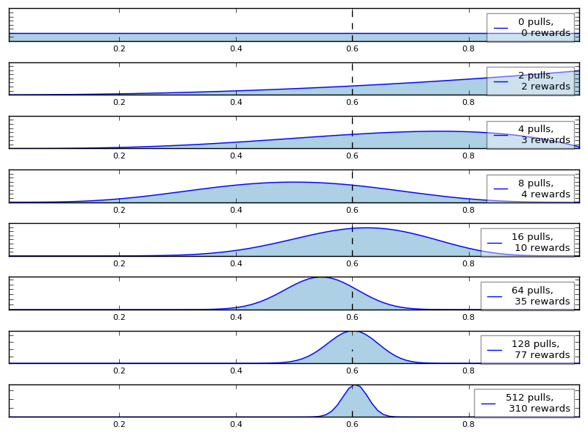}
\caption{Exemplary evolution of the belief distribution wrt. $\mu$ for a Bernoulli bandit. See text for details.}
\label{fig:beta}
\end{figure}

Figure \ref{fig:beta} shows the evolution of the belief distribution for an example Bernoulli bandit with a probability $\mu = 0.6$ of providing a reward when pulling the bandit. We want to infer $\mu$ by sampling from the bandit, where each pull either yields a reward or no reward. Initially, no data is observed and the prior is uniform, representing that all values for $\mu$ are equally likely a prior (before seeing any data). The images show the evolution of the belief distribution for an increasing number of pulls and correspondingly observed rewards from top to bottom. An increasing number of pulls yields a sharper (i.e. less uncertain) estimate of $\mu$. The probability mass (i.e. the blue integral) to the right of any value for $\mu$ is the confidence that the true $\mu$ lies above the value. For example, the integral to the right of $0.6$ denotes the confidence that $\mu$ is indeed larger than $0.6$. 

We can think of the environment in combination with a particular policy as our Bernoulli bandit, providing a reward when given constraints are satisfied. The agent wants to infer the satisfaction probability, and can quantify its confidence in the result using Bayesian modeling as outlined above. 

Algorithm \ref{alg:BV} shows the pseudo code for Bayesian verification (BV) of a policy that is parametrized with parameters $\theta$. It repeatedly generates an episode (line 4) and computes its cumulative cost (line 5), evaluates whether it satisfies the safety requirement by testing whether the episode cost equals zero (line 6), and updates its confidence in satisfaction accordingly (line 7). BV terminates when the confidence reaches a given bound (lines 8 and 9).

\begin{algorithm}[H]
	\begin{algorithmic}[1]
		\Procedure {BV}{Policy parameters $\theta$, CMDP $m \in M$, requirement $\mathbb{P}_{\geq p_\mathrm{req}}(\varphi) \textrm{ and } \mathbb{C}_{\geq c_\mathrm{req}}$}
		\State $s, v \gets 0$
		\Loop
		\State $\vec{e} \sim m(\theta)$
		\State compute $\mathcal{C}_\varphi$ wrt. $\vec{e}$
		\State update $s$ or $v$ wrt. $\mathcal{C}_\varphi =_? 0$
		\State determine $c_\mathrm{sat}$ wrt. Eq. \ref{eq:c_sat}
		\If {$c_\mathrm{sat} \geq c_\mathrm{req}$}
		\Return true
		\EndIf
		\If {$1 - c_\mathrm{sat} \geq c_\mathrm{req}$}
		\Return false
		\EndIf
		\EndLoop
		\EndProcedure
	\end{algorithmic}
	\caption{Bayesian verification}
	\label{alg:BV}
\end{algorithm}

While this approach allows to verify a given policy after training, it does not directly provide a way to synthesize a policy that is likely to be verified. We will provide an approach to this problem in the next section.

\section{Safe Neural Evolutionary Strategies}

The previous section outlined a methodology for engineering safe policy synthesis based on specification as a CMDP, safe RL and Bayesian verification. In this section, we propose Safe Neural Evolutionary Strategies (SNES) for learning policies that are likely to be positively verified. SNES weights return and cost in the process of policy synthesis based on Bayesian confidence estimates obtained in the course of learning.

\subsection{Evolutionary Strategies}

Evolutionary strategies (ES) \cite{salimans2017evolution} is a gradient free, search-based optimization algorithm that has shown competitive performance in reinforcement learning tasks using deep learning for function approximation. ES is attractive as it is not based on backpropagation and can therefore be parallelized straightforwardly. Also, it does not require expensive GPU hardware for efficient computation. 

The basic ES procedure is shown in Algorithm \ref{alg:ES}. ES works by maintaining the parameters $\theta$ of the current solution. It then generates $N \in \mathbb{N}^+$ slightly perturbed offspring from this solution to be evaluated on the optimization task $f$, for example optimizing expected episode return of a policy in an MDP (lines 4 to 7). In our case, we use $N$ normally distributed samples with a mean of $0$ and a standard deviation of $\sigma$. 
The current solution is then updated by moving the solution parameters into the direction of offspring weighted by their respective return, in expectation increasing effectiveness of the solution (lines 8 and 9).

We normalize a set of values $X \in \mathbb{R}^{\mathbb{N}^+}$ to zero mean and unit standard deviation with the following normalization procedure.
\begin{equation}
    \mathrm{normalize}(X) =_\mathrm{def} \forall x \in X : x \leftarrow \dfrac{x - \mathrm{mean}(X)}{\mathrm{std}(X)}
\end{equation}

\begin{algorithm}[H]
	\begin{algorithmic}[1]
		\Procedure {ES}{population size $N \in \mathbb{N}^+$, perturbation rate $\sigma \in \mathbb{R}^+$, learning rate $\alpha \in (0, 1]$, task $f : \Theta \rightarrow \mathbb{R}$}
		\State initialize parameters $\theta$
		\While {learning}
		
		\For{$i \in \{1, ..., N\}$}
		
		\State $\epsilon_i \sim \mathrm{Normal}(0, \sigma, N)$
		\Comment{perturb offspring}
		\State $\theta_i \leftarrow \theta + \epsilon_i$
		
		\State $\mathcal{R}_i \leftarrow f(\theta_i)$
		\Comment{determine return}
		
		\EndFor
		
		\State $\mathrm{normalize}(\bigcup_i \mathcal{R}_i)$
		
		\Comment{update solution}
		\State $\theta \leftarrow \theta + \frac{\alpha}{\sigma N} \sum_{j = 1}^{N} \mathcal{R}_j * \epsilon_j$
		
		\EndWhile
		\EndProcedure
	\end{algorithmic}
	\caption{ES}
	\label{alg:ES}
\end{algorithm}

\subsection{Safe Neural Evolutionary Strategies}

Safe Neural Evolutionary Strategies (SNES) combines ES with Bayesian verification in the learning process to adaptively weight return and cost in the course of policy synthesis such that the resulting policy is likely to be positively verified.

SNES is an instance of safe RL based on basic ES for parameter optimization,  additionally performs a Bayesian verification step in each iteration and uses the current verification confidence for balancing return maximization and constraint satisfaction. The resulting confidence estimate in positive verification is then used to determine the weighting $\lambda$ of return and cost.

In order to account for confidence requirements, the confidence estimate $c_\mathrm{sat}$ from Bayesian verification is set in relation to the required confidence $c_\mathrm{req}$ such that if confidence in constraint satisfaction is lower than required, only costs are reduced in the parameter update. If the constraint is satisfied with enough confidence, the influence of return is gradually increased.
\begin{equation}
    \lambda \leftarrow \dfrac{\max(0, c_\mathrm{sat} - c_\mathrm{req})}{1 - c_\mathrm{req}}
    \label{eq:lambda}
\end{equation}



SNES is shown in Algorithm \ref{alg:SNES}. SNES generates offspring by adding noise to the parameters of the current policy (lines 7 and 8). It evaluates the offspring by generating an episode from the MDP (line 9) and computes its return and cost (line 10), and whether it satisfies or violates the requirement (line 11). It uses this information to update the Lagrangian of the constrained optimization problem to weight return and cost in the learning process (lines 12 to 14). Learning is done by updating parameters of a policy, weighted accordingly to normalized return and cost (lines 15 to 18).

\begin{algorithm}[H]
	\begin{algorithmic}[1]
		\Procedure {SNES}{population size $N \in \mathbb{N}^+$, perturbation rate $\sigma \in \mathbb{R}^+$, learning rate $\alpha \in (0, 1]$, CMDP $m \in M$, requirement $\mathbb{P}_{\geq p_\mathrm{req}}(\varphi) \textrm{ and } \mathbb{C}_{\geq c_\mathrm{req}}$}
		\State initialize parameters $\theta$
		\State $s, v \leftarrow 0$
		\Comment{initally no satisfying or violating episodes}
		\State $\lambda \leftarrow 1$
		\While {learning}
		  
		  \For{$i \in \{1, ..., N\}$}
		  
		  \State $\epsilon_i \sim \mathrm{Normal}(0, \sigma, N)$
		  \Comment{perturb current parameters}
		  
		  \State $\theta_i \leftarrow \theta + \epsilon_i$
		  
		  \State $\vec{e} \sim m(\theta_i)$
		  \Comment{evaluate perturbation}
		  \State compute $\mathcal{R}_i, \mathcal{C}_{\varphi, i}$ wrt. $\vec{e}$

		  \State $s \leftarrow s + \mathbb{I} ( \mathcal{C}_{\varphi, i} = 0 )$
		  \Comment{count satisfying episodes}
		  
		  \EndFor
		  
		  \State $v \leftarrow v + N - s$
		  \Comment{number of violating episodes}
		  
		  
		  \State $c_\mathrm{sat} \leftarrow 1 - \mathrm{Beta}(s + 1, v + 1).\mathrm{cdf}(p_\mathrm{req})$
		  \Comment{determine confidence}
		  
		  \State $\lambda \leftarrow \dfrac{\max(0, c_\mathrm{sat} - c_\mathrm{req})}{1 - c_\mathrm{req}}$
		  \Comment{set $\lambda$ wrt. requirement}

		  \State $\mathrm{normalize}(\bigcup_i \mathcal{R}_i)$
		  \Comment{normalize return and cost}
		  \State $\mathrm{normalize}(\bigcup_i \mathcal{C}_{\varphi, i})$
		  
		  \State $\theta \leftarrow \theta + \frac{\alpha \lambda}{\sigma N} \sum_{j = 0}^{N} \mathcal{R}_j * \epsilon_j$
		  \Comment{update parameters}
		  \State $\theta \leftarrow \theta - \frac{\alpha (1 - \lambda)}{\sigma N} \sum_{j = 0}^{N} \mathcal{C}_{\varphi, j} * \epsilon_j$
		  
		\EndWhile
		\EndProcedure
	\end{algorithmic}
	\caption{Safe Neural Evolutionary Strategies (SNES) for policy synthesis under probabilistic constraints}
	\label{alg:SNES}
\end{algorithm}

\paragraph{Remark: SNES vs. maximum likelihood calibration of the Langrangian}
To show the effect of the Bayesian treatment and the necessity of computing confidence for adapting $\lambda$ in the learning process, we compared SNES to a naive variant for tuning $\lambda$ (for results cf. Section \ref{par:SNES_vs_MLE}). Here, we use a maximum likelihood estimate $\hat p_\mathrm{sat}$ for satisfaction probability and adjust $\lambda$ according to the following rule, replacing lines 13 and 14 in Algorithm \ref{alg:SNES}.
\begin{align}
\hat p_\mathrm{sat} &\leftarrow \dfrac{s}{N} \\
\lambda &\leftarrow \dfrac{\max(0, \hat p_\mathrm{sat} - p_\mathrm{req})}{1 - p_\mathrm{req}}
\label{eq:MLE}
\end{align}
Note that the naive approach does not account for confidence in its result.

\section{Experiments}

In this section, we report our empirical results obtained when evaluating SNES for two domains:\footnote{Code for our experiments is available at \url{https://github.com/lenzbelzner/psyco}.}
\begin{itemize}
\item The \textit{Particle Dance} domain has continuous state and action spaces, and a fixed episode length. Its initial policy is likely to satisfy given constraints when starting the learning process.
\item The \textit{Obstacle Run} domain has discrete state and actions spaces, and has an adaptive episode length based on a termination criterion. Here, the initial policies are likely to violate given constraints.
\end{itemize}

For each of the domains, we describe the setup of our experiment by presenting the respective domain, the rewards, costs, and requirements, the neural network for modeling the policy, and the parameters for performing the experiments.
Then we analyze the episode return, satisfaction probability and confidence, and show the results of Bayesian verification for both domains. For the \textit{Particle Dance} domain, we additionally compare the SNES calibration of the Langrangian with a maximum likelihood approach.

\subsection{Particle Dance} \label{particle dance}


In the \textit{Particle Dance} domain, an agent has to learn to follow a randomly moving particle as closely as possible while keeping a safe distance. The \textit{Particle Dance} domain is continuous with fixed episode length, and an initial policy is likely to satisfy given constraints.

\subsubsection{Setup}
Agent and particle have a position $x \in [-2, 2]^2$ and a velocity $v \in [-0.1, 0.1]^2$. The state space $S$ describes the positions and velocities of both agent and particle. We restrict positions and velocities to their respective boundaries by clipping any exceeding values. The state also keeps count of agent-particle collisions $n_c \in \mathbb{N}$ (see below for their definition). Note that the collision counter is not observed by the agent in our setup, i.e. is not used as input for training and querying its policy.
\begin{equation}
    S : [-2, 2]^4 \times [-0.1, 0.1]^4 \times \mathbb{N}
\end{equation}
The initial positions are sampled from $[-1, 1]^4$ uniformly at random. The initial velocities are fixed to zero, as is the initial collision counter.
\begin{equation}
     \rho : \mathcal{U}([-1, 1]^4) \times [0, 0]^4 \times 0
\end{equation}
The agent can choose its acceleration at each time step. This yields the continuous action space A.
\begin{equation}
    A : [-0.1, 0.1]^2
\end{equation}
The agent is accelerated at each time step by a value $a \in [-0.1, 0.1]^2$. The particles' acceleration is sampled uniformly at random at each time step.
Positions are updated wrt. current velocities. We define a collision radius $d_\mathrm{min} \in \mathbb{R}^+$ and induce a cost when the agent is closer to the particle than this radius and update $n_c$ accordingly. Let $s \in S$ be the systems current state and $a \in A$ be the action executed by the agent, then the transition distribution $T : p(S \vert S, A)$ is given by the following sequence of assignments:
\begin{align*}
    & s = (x_\mathrm{agent}, x_\mathrm{particle}, v_\mathrm{agent}, v_\mathrm{particle}, n_c)\\
    & T(s, a) \sim
    \begin{cases}
        v_\mathrm{particle} \leftarrow v_\mathrm{particle} + \mathcal{U}[-0.1, 0.1]^2 \\
        x_\mathrm{particle} \leftarrow x_\mathrm{particle} + v_\mathrm{particle} \\
        v_\mathrm{agent} \leftarrow v_\mathrm{agent} + a \\
        x_\mathrm{agent} \leftarrow x_\mathrm{agent} + v_\mathrm{agent} \\
        n_c \leftarrow n_c + \mathbb{I}(d(x_\mathrm{agent}, x_\mathrm{particle}) < d_\mathrm{min})
    \end{cases}
\end{align*}

Note that $T$ is not known by the agent and detailed here only to render our experimental setup reproducible.

\paragraph{Reward and Safety Predicate}
The agent gets a reward at each step of an episode motivating it to get as close to the particle as possible. We define a collision radius $d_\mathrm{min} \in \mathbb{R}^+$ and induce a cost when the agent is closer to the particle than this radius. We also define a collision counter $n_c \in \mathbb{N}$ as an atomic proposition of states which allows us to specify a number of collisions $n_{\max}$. Note that $n_c$ is not observed by the agent in our experiments. We set $n_{\max} \in {1, 4}$ and $d_\mathrm{min} = 0.1$ in our experiments. Let $s' = (x'_\mathrm{agent}, x'_\mathrm{particle}, v'_\mathrm{agent}, v'_\mathrm{particle}, n'_c)$.
\begin{equation}
R(s, a, s') = - d(x'_\mathrm{agent}, x'_\mathrm{particle})
\end{equation}
\begin{equation}
    \phi = d(x'_\mathrm{agent}, x'_\mathrm{particle}) \geq d_\mathrm{min} \vee n'_c \leq n_{\max}
\end{equation}

\paragraph{Requirements and Cost}
The reward computes the negative distance between particle and agent. Minimizing the distance means maximizing the reward. Thus the optimizing goal is to maximize the expectation of the return $\mathcal{R}$ (see  (\ref{eq:optimize})): 
\begin{equation}
\mathbf{Goal} \text{ Optimize } \mathit{Return} : \max \mathbb{E}(\mathcal{R})
\end{equation}

The safety constraint requires the agent to keep a minimum distance of the particle except in $n_{max}$ cases. We set the required probability for satisfying the constraint $p_\mathrm{req} = 0.85$ and the required confidence $c_\mathrm{req} = 0.98$.
\begin{equation}
\mathbf{Goal} \textrm{ Constraint } \mathit{Bounded Collisions} :\mathbb{P}_{\geq 0.85}(\square \phi) \textrm{ and } \mathbb{C}_{\geq 0.98}
\end{equation}
The cost induced by the safety constraint is given by equation ~\ref{eq:formula1}.
\paragraph{Policy Network}
We model the policy of our agents as a feedforward neural network with parameters $\theta$. Our network consists of an input layer with dimension 8 (position and velocity of agent and particle), a hidden layer with dimension 32, and has an output dimension of 2 (two dimensions of acceleration). Let $\theta = \{\theta_1, \theta_2\}$ be the networks' weights in the input and hidden layer respectively, and let $f_1$ and $f_2$ be non-linear activation functions, with $f_1$ being a rectified linear unit \cite{glorot2011deep} and $f_2$ being \textit{tanh} in our case. Then, the networks output is given as follows.
\begin{equation} \label{eq:output}
    y \leftarrow f_2(\theta_2 f_1(\theta_1 x + 1) + 1)
\end{equation}

\paragraph{Other Parameters}
We report our results for  episode length $n = 50$, population size $N = 20$, learning rate $\alpha = .01$ and perturbation rate $\sigma = .1$. Experiments with other parameters yielded similar results.

\paragraph{Experimental Setup}
Each experimental run comprised learning a policy with SNES over 60000 episodes. Every 1000 episodes, we performed Bayesian verification for a maximum of 1000 episodes (outside the learning loop of SNES) to evaluate the policy synthesized by SNES up to that point.

We repeated the experiment five times and show mean values as solid lines and standard deviation by shaded areas in our figures.

\paragraph{A Note on Performance}
The performance of SNES depends on the neural network architecture, the simulation of the MDP transitions and the used hardware. In our setup using a laptop computer, each step took less than a millisecond of execution time.

\subsubsection{Results}

\begin{figure}
\centering
\includegraphics[width=0.8\textwidth]{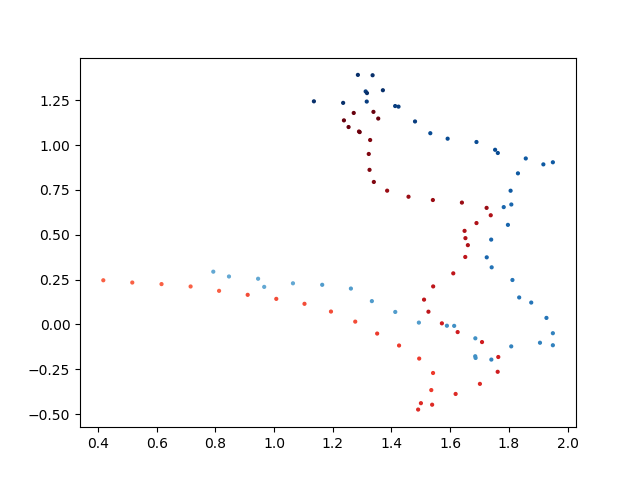}
\caption{Sample trajectories of the particle (light to dark blue, color gradient denotes time) and the agent (light to dark red). The shown trajectory has been learned by an unconstrained agent and achieves a return of ca. $-15$ without incurring a collision.}
\label{fig:follow_1}
\end{figure}

The SNES agent learns to follow the particle closely. In Figure \ref{fig:follow_1} we see sample trajectories of the particle and the agent (color gradients denote time). We observe that the synthesized policy learned the task to follow the particle successfully.

\begin{figure}
\centering
\includegraphics[width=.8\textwidth]{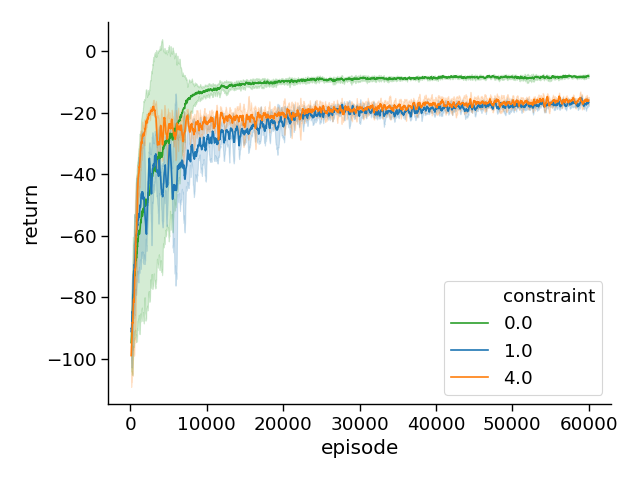}
\caption{Episode return for various constraints. Constraint 0.0 denotes unconstrained policy synthesis.}
\label{fig:reward}
\end{figure}

We can observe the effect of SNES learning unconstrained and constrained policies on the obtained episode return in Figure \ref{fig:reward}. Return and constraint define a Pareto front in our domain: Strengthening the constraint reduces the space of feasible policies, and also reduces the optimal return that is achievable by a policy due to increased necessary caution when optimizing the goal, i.e. when learning to follow the particle as close as possible.

\begin{figure}
\centering
 \includegraphics[width=.8\textwidth]{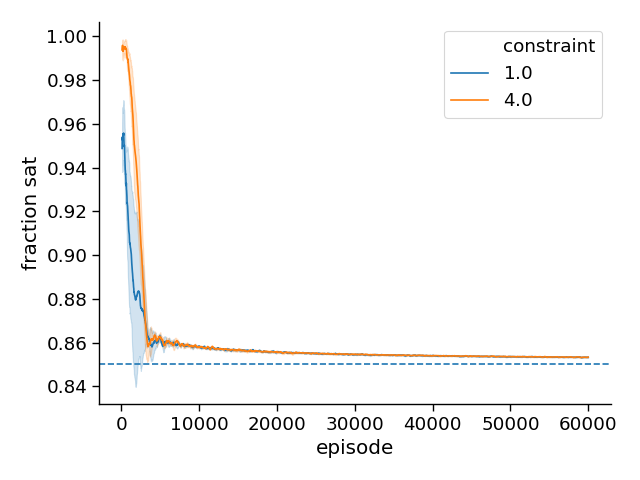}
 \caption{Proportion of episodes satisfying cost requirement.}
 \label{fig:n_sat}
\end{figure}

Figure \ref{fig:n_sat} shows the proportion of episodes that satisfy the given requirement. We can see that the proportion closely reaches the defined bound, shown by the dashed vertical line. Note that the satisfying proportion is closely above the required bound.

\begin{figure}
\centering
\includegraphics[width=.8\textwidth]{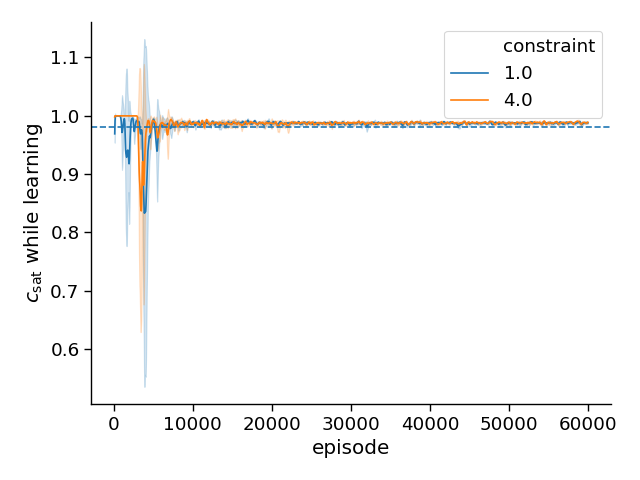}
\caption{Confidence $c_\mathrm{sat}$ in satisfying specification based on observations in the course of learning.}
\label{fig:c_sat}
\end{figure}

Figure \ref{fig:c_sat} shows the confidence of the learning agent in its ability to satisfy the given requirement based on the observations made in the learning process so far. The confidence is determined from the Beta distribution maintained by SNES over the course of training. Note that the confidence is mostly kept above the confidence requirement given in the specification. This shows SNES is effectively incorporating observations, confidence and requirement into its learning process.

\begin{figure}
\centering
\includegraphics[width=.8\textwidth]{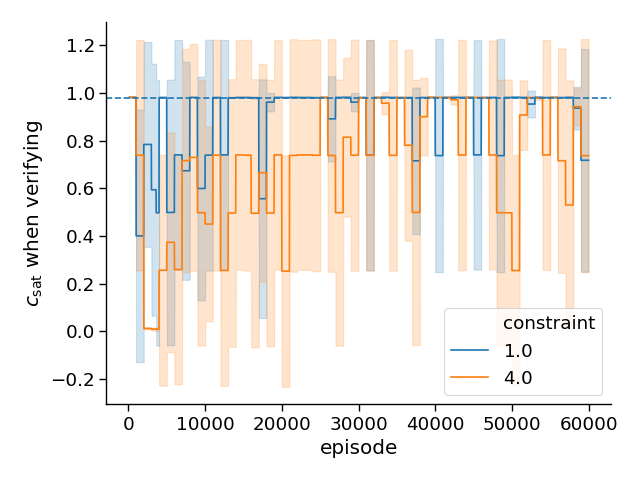}
\caption{Confidence obtained when exhaustively verifying stationary current policies over the course of learning with Bayesian verification BV every 1000 episodes.}
\label{fig:verifyability}
\end{figure}

Figure \ref{fig:verifyability} shows the results of Bayesian verification (see Alg.~\ref{alg:BV}) performed throughout the learning process every 1000 episodes. Here, we fix the current policy, and perform Bayesian verification of the policy wrt. the given requirement. The quantity measured is the confidence in requirement satisfaction after either surpassing $c_\mathrm{req}$ (i.e. the policy satisfies the requirement with high confidence), falling below $1 - c_\mathrm{req}$ (i.e. the policy violates the requirement with high confidence) or after a maximum of 1000 verification episodes. We can see that the confidence in having learned a policy that satisfies the requirement is increasing over the course of training. However, as the goals of optimizing return and satisfying collision constraints are contradictory in the Particle Dance domain, SNES may still produce policies that eventually violate the specification.

\begin{figure}
\centering
\includegraphics[width=\textwidth]{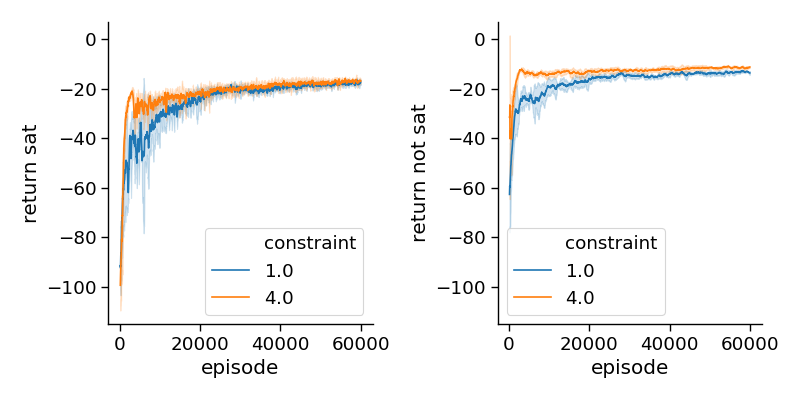}
\caption{Return of episodes satisfying (left) and violating (right) constraints.}
\label{fig:r_sat}
\end{figure}

\begin{figure}
\centering
\includegraphics[width=\textwidth]{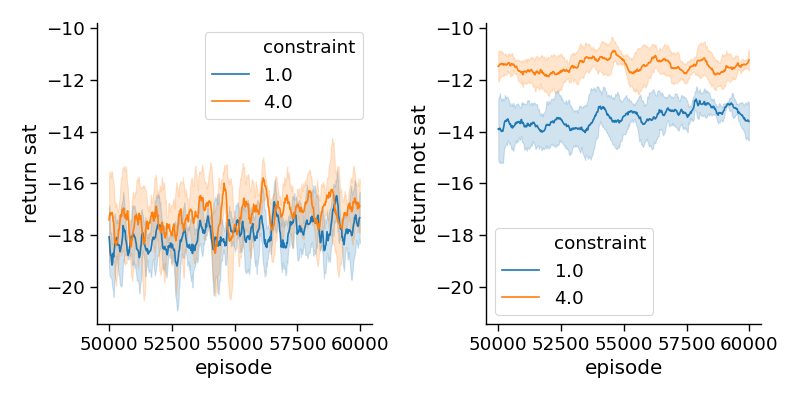}
\caption{Return of episodes satisfying (left) and violating (right) constraints as shown in Figure \ref{fig:r_sat}, showing the final episodes in more detail. The less constrained policy is able to optimize its return more effectively.}
\label{fig:r_sat_zoom}
\end{figure}

\begin{figure}
\centering
\includegraphics[width=\textwidth]{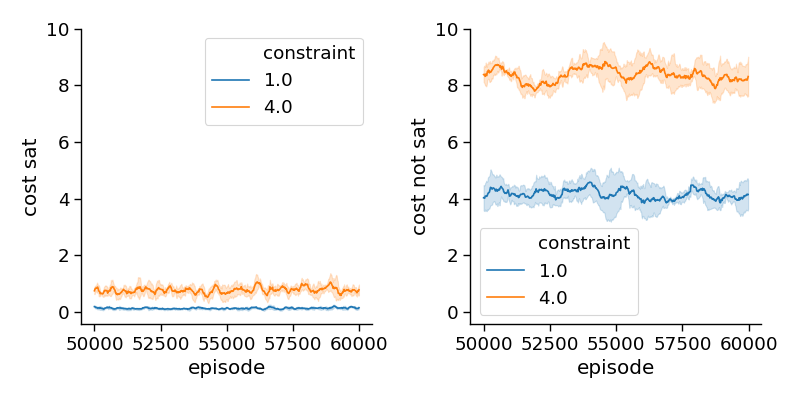}
\caption{Number of collision events (i.e. cost) of episodes satisfying (left) vs. violating (right) constraints.}
\label{fig:e_sat}
\end{figure}

Figures \ref{fig:r_sat}, \ref{fig:r_sat_zoom} and \ref{fig:e_sat} show return and collisions (i.e. cost) obtained, split by episodes that satisfy the constraint and those that violate it. We can see the violating episodes are more effective in terms of return but keep collisions well below the requirement, highlighting again the Pareto front of return and cost given by our domain. Note that we smooth the shown quantity over the last 100 episodes, and that collision only can take discrete values. This may explain that the shown quantity is well below the theoretically given boundaries (one or four in our case). SNES is able to learn policies that exploit return in the defined proportion of episodes, and to optimize wrt. the Pareto front of return and cost otherwise.

\paragraph{Results on SNES vs. maximum likelihood calibration of the Lagrangian}
\label{par:SNES_vs_MLE}

\begin{figure}
	\centering
	\includegraphics[width=\textwidth]{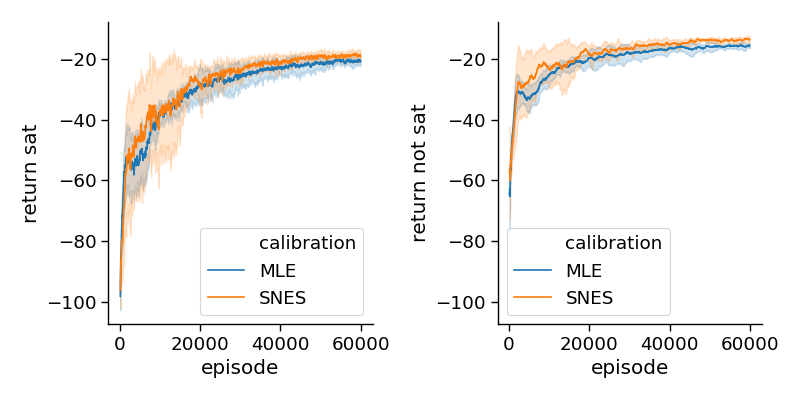}
	\caption{Return of episodes satisfying (left) and violating (right) constraints for SNES and MLE of $\hat p_\mathrm{sat}$.}
	\label{fig:r_sat_vs_naive}
\end{figure}

\begin{figure}
	\centering
	\includegraphics[width=\textwidth]{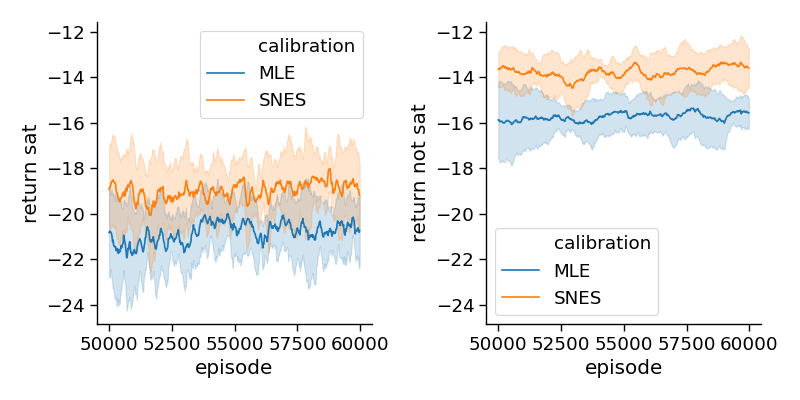}
	\caption{Return of episodes satisfying (left) and violating (right) constraints for SNES and MLE of $\hat p_\mathrm{sat}$ as shown in Figure \ref{fig:r_sat_vs_naive}, showing the final episodes in more detail. SNES is able to exploit the Pareto front of optimization and constraints more effectively.}
	\label{fig:r_sat_vs_naive_zoom}
\end{figure}

\begin{figure}
	\centering
	\includegraphics[width=.8\textwidth]{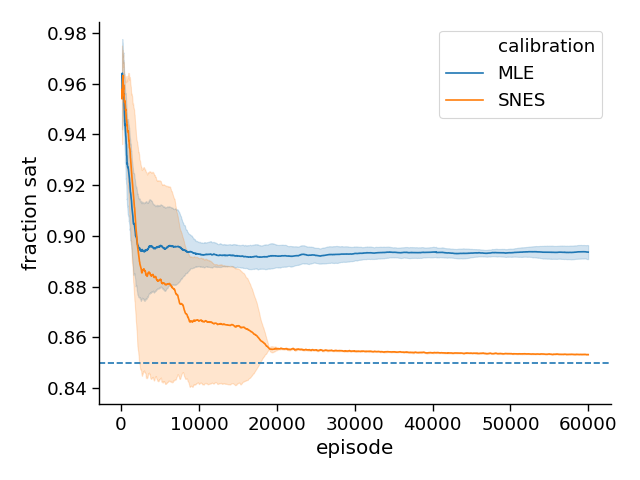}
	\caption{Proportion of satisfying episodes for SNES and MLE of $\hat p_\mathrm{sat}$.}
	\label{fig:n_sat_vs_naive}
\end{figure}

We compared SNES' Bayesian approach to calibrating $\lambda$ (Eq. \ref{eq:lambda}) to a maximum likelihood variant (Eq. \ref{eq:MLE}). Figures \ref{fig:r_sat_vs_naive}, \ref{fig:r_sat_vs_naive_zoom} and \ref{fig:n_sat_vs_naive} show return and cost for both approaches, separated by satisfying and violating episodes. We show the results for $n_{\max} = 1$, aggregated for 12 repetitions of the experiment.

The MLE approach over-satisfies the given constraint, and consequently does not yield as high returns as SNES. In contrast, SNES is able to find a local Pareto optimum of return and constraint satisfaction.

\subsection{Obstacle Run} \label{obstacle run}

In the \textit{Obstacle Run} domain, an agent has to reach a target position while not colliding with a moving obstacle. In contrast to \textit{Particle Dance}, \textit{Obstacle Run} has discrete states and actions, and episodes terminate when the agent reaches the target position.
Also, an initial policy is likely to violate the given constraints when starting the learning process.

\subsubsection{Setup}
The setup is similar to the Particle Dance, except that positions are discrete with $x \in \{0, ..., 4\}^2$ and that the agent does not change the velocity but only the movement direction in each step. 
The state space $ S : \{0, ..., 4\}^4 \times \mathbb{N}$ consists of the positions of agent and obstacle and of a collision counter $n_c \in \mathbb{N}$. As in Particle Dance, the collision counter is not used as input for training and querying the agent’s policy  and positions are restricted to their respective boundaries by clipping any exceeding values.

%
The initial positions are sampled from $\{0, ..., 4\}^4$ uniformly at random. The initial collision counter is set to zero.
\begin{equation}
     \rho : \mathcal{U}(\{0, ..., 4\}^4) \times 0
\end{equation}

The agent can choose its movement direction at each time step. This yields the discrete action space A.
\begin{equation}
    A : \{(0, 0), (1, 0), (0, 1), (-1, 0), (0, -1)\}
\end{equation}
The agent is moves at each time step by according to its chosen action. The obstacle's movement is sampled uniformly at random from $A$ at each time step.
Positions are updated wrt. actions. A collision occurs if agent and obstacle share the same position, and update $n_c$ accordingly. Let $s \in S$ be the systems current state and $a \in A$ be the action executed by the agent, then the transition distribution $T : p(S \vert S, A)$ is given by:
\begin{align*}
    & s = (x_\mathrm{agent}, x_\mathrm{obstacle}, n_c)\\
    & T(s, a) \sim
    \begin{cases}
        x_\mathrm{obstacle} \leftarrow x_\mathrm{obstacle} + \mathcal{U}(A) \\
        x_\mathrm{agent} \leftarrow x_\mathrm{agent} + a \\
        n_c \leftarrow n_c + \mathbb{I}(x_\mathrm{agent} = x_\mathrm{obstacle})
    \end{cases}
\end{align*}

Note that $T$ is not known by the agent and detailed here only to render our experimental setup reproducible.

We fix the target position $x_\mathrm{target} = (0, 0)$ and end an episode if $x_\mathrm{agent} = x_\mathrm{target}$.

\paragraph{Requirements, Reward, and Cost}
The agent gets a reward of $-1$ at each step of an episode motivating it to reach the target as fast as possible. The optimizing goal is to maximize the expectation of the return $\mathcal{R}$ (see  (\ref{eq:optimize})). Note that this is achieved by reaching the target position as fast as possible.
\begin{equation}
R(s, a, s') = -1
\end{equation}
%
\begin{equation}
\mathbf{Goal} \text{ Optimize } \mathit{Return} : \max \mathbb{E}(\mathcal{R})
\end{equation}

The safety constraint requires the agent to avoid the (moving) obstacle except in $n_{max}$ cases. We set the required probability for satisfying the constraint $p_\mathrm{req} = 0.9$ and the required confidence $c_\mathrm{req} = 0.98$.
\begin{equation}
    \phi = x_\mathrm{agent} \not= x_\mathrm{target} \vee n_c \leq n_{\max}
\end{equation}
\begin{equation}
\mathbf{Goal} \textrm{ Constraint } \mathit{Bounded Catches} :\mathbb{P}_{\geq 0.9}(\square \phi) \textrm{ and } \mathbb{C}_{\geq 0.98}
\end{equation}
The cumulative cost is again given by equation (\ref{eq:formula1}). In our experiments we set $n_{\max} \in {1, 4}$ . 

\paragraph{Policy Network}
As in Particle Dance, the policy network is a feedforward neural network with a hidden layer of dimension 32 and its output is computed by equation (\ref{eq:output}). The input layer has dimension 4 (position of agent and obstacle) and the output layer has dimension 5.

Actions are chosen by identifying the index of the maximum output and choosing one of the five available actions accordingly.

\paragraph{Other Parameters}
As for Particle Dance we report our results for a maximum episode length $n = 50$, population size $N = 20$, learning rate $\alpha = .01$ and perturbation rate $\sigma = .1$. Experiments with other parameters yielded similar results.

\paragraph{Experimental Setup}
Each experimental run comprised learning a policy with SNES over 20000 episodes. Every 1000 episodes, we performed Bayesian verification for a maximum of 1000 episodes (outside the learning loop of SNES) to evaluate the policy synthesized by SNES up to that point.

We repeated the experiment five times and show mean values as solid lines and standard deviation by shaded areas in our figures.

\subsubsection{Results}

\begin{figure}[htb]
    \centering
    \begin{minipage}[t]{0.49\textwidth}
        \centering
        \includegraphics[width=\textwidth]{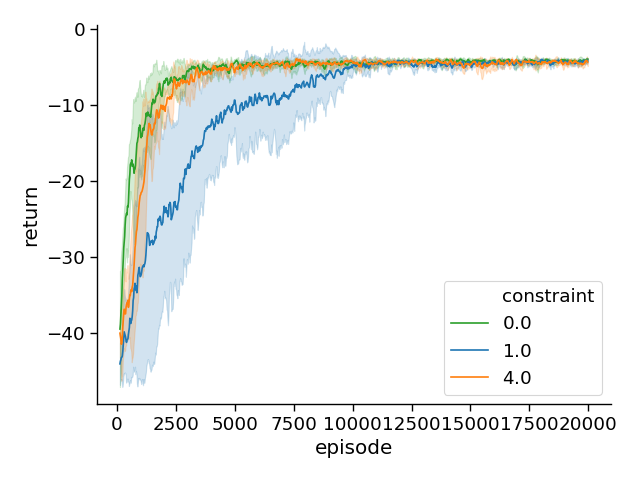}
        \caption{ \textit{Obstacle Run}: Episode return for various constraints.}
       \label{fig:reward_2}
    \end{minipage}
    \hfill
    \begin{minipage}[t]{0.49\textwidth}
        \centering
        \includegraphics[width=\textwidth]{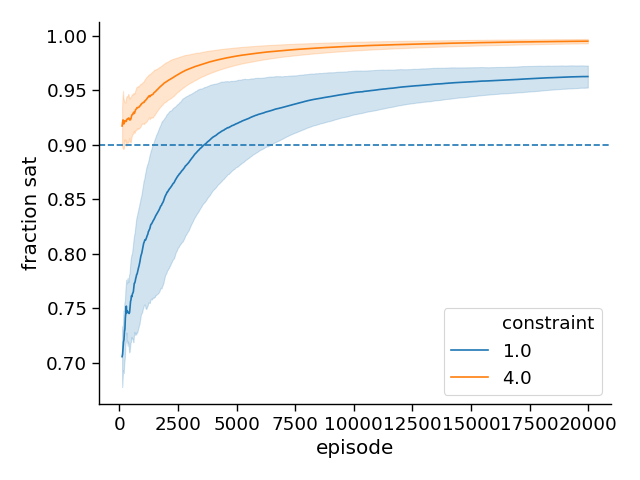}
        \caption{ \textit{Obstacle Run}: Proportion of episodes satisfying cost requirement.}
       \label{fig:n_sat_2}
    \end{minipage}
\end{figure}

We can observe the effect of SNES learning unconstrained and constrained policies on the obtained episode return in Figure \ref{fig:reward_2}. Constraint 0.0 denotes unconstrained policy synthesis. Return and constraint define a Pareto front in our domain: Strengthening the constraint reduces the space of feasible policies, and also reduces the optimal return that is achievable by a policy due to increased necessary caution when optimizing the goal, i.e. when learning to avoid the obstacle as good as possible.
%
%

Figure \ref{fig:n_sat_2} shows the proportion of episodes that satisfy the given requirement. We can see that the easily surpasses defined bound, shown by the dashed vertical line. In contrast to \textit{Particle Dance}, the agent is able to find policies that easily satisfy the given constraint by reaching the target as fast as possible, thus reducing the probability of a collision.
%

\begin{figure}[htb]
    \centering
    \begin{minipage}[t]{0.49\textwidth}
        \centering
        \includegraphics[width=\textwidth]{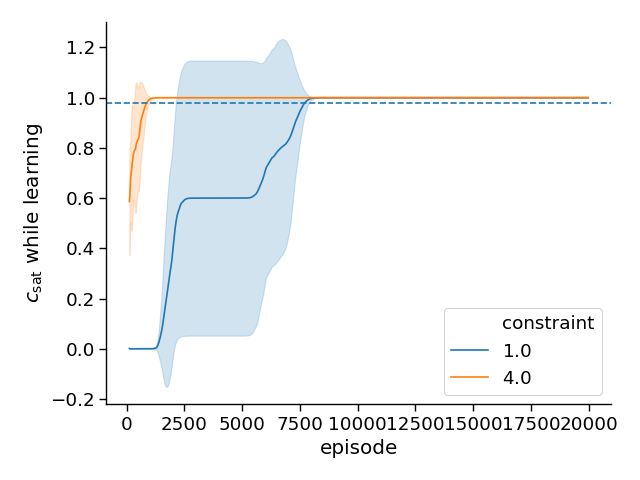}
        \caption{ \textit{Obstacle Run}:  Confidence $c_\mathrm{sat}$ in satisfying specification based on observations in the course of learning.}
       \label{fig:c_sat_2}
    \end{minipage}
    \hfill
    \begin{minipage}[t]{0.49\textwidth}
        \centering
        \includegraphics[width=\textwidth]{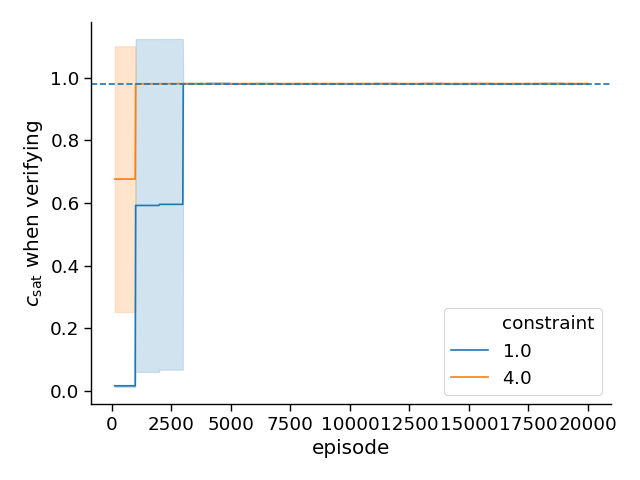}
        \caption{ \textit{Obstacle Run}: Confidence obtained when exhaustively verifying stationary current policies over the course of learning with Bayesian verification BV every 1000 episodes.}
       \label{fig:verifyability_2}
    \end{minipage}
\end{figure}

Figure \ref{fig:c_sat_2} shows the confidence of the learning agent in its ability to satisfy the given requirement based on the observations made in the learning process so far. The confidence is determined from the Beta distribution maintained by SNES over the course of training.
The agent is able to learn a policy that allows to both optimize return and satisfy the given constraint in a stable way.

Figure \ref{fig:verifyability_2} shows the results of Bayesian verification performed throughout the learning process every 1000 episodes. In contrast to the Particle Dance domain, in the Obstacle Run domain the agent is likely to violate the constraint at the start of the learning process. Therefore, in the beginning it tends to underestimate its ability to satisfy the constraint, as can be seen from comparing its confidence while learning (Figure \ref{fig:c_sat_2}) with the (offline) confidence acquired from verifying (Figure \ref{fig:verifyability_2}).

%

\begin{figure}
\centering
\includegraphics[width=\textwidth]{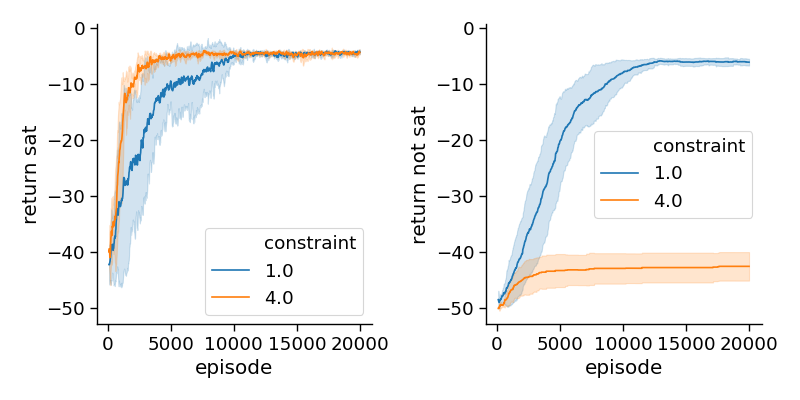}
\caption{\textit{Obstacle Run}: Return of episodes satisfying (left) and violating (right) constraints.}
\label{fig:r_sat_2}
\end{figure}

\begin{figure}
\centering
\includegraphics[width=\textwidth]{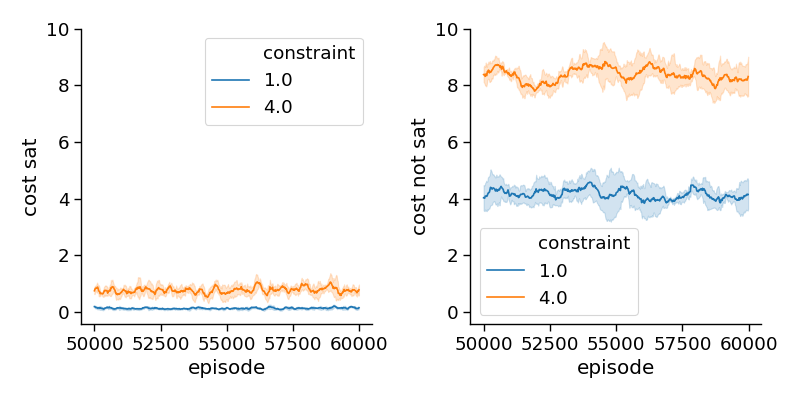}
\caption{\textit{Obstacle Run}: Number of collision events (i.e. cost) of episodes satisfying (left) vs. violating (right) constraints.}
\label{fig:e_sat_2}
\end{figure}

Figures \ref{fig:r_sat_2} and \ref{fig:e_sat_2} show return and collisions (i.e. cost), split by episodes that satisfy the constraint and those that violate it. We can see the violating episodes are more effective in terms of return but keep collisions well below the requirement, highlighting again the Pareto front of return and cost given by our domain. SNES is able to learn policies that exploit return in the defined proportion of episodes, and to optimize wrt. the Pareto front of return and cost otherwise.

\section{Related Work}

For synthesizing policies of autonomous and adaptive systems, PSyCo comprises  a systematic development method, algorithms for  safe and robust reinforcement learning, and a Bayesian verification method which is related to MDP model checking and statistical model checking approaches. In this section we discuss related work in these areas.  

\paragraph{Systematic Development of Adaptive Systems}
PSyCo borrows its notion of goals from KAOS~\cite{DardenneLF93}, an early method for goal-oriented requirements engineering.
KAOS distinguishes hard and soft goals, is formally based on linear temporal logic, and proposes activities for refining the goals and deriving operation requirements which serve as the basis for system design. In contrast to PSyCo, it does neither cover system design nor implementation.

SOTA~\cite{SOTA20}  is a modern requirements engineering method for autonomous and collective adaptive systems with a specific format for goals.  Properties of goals can be analyzed by a modelchecking tool~\cite{AbeywickramaZ12} based on  LTL formulas and the  LTSA modelchecker~\cite{MK06}. SOTA does neither address system design nor implementation; but it was used for requirements specification in the systematic construction process for autonomous ensembles~\cite{WirsingHTZ11} of the ASCENS project~\cite{ASCENS15}. 

The ensemble development life cycle EDLC ~\cite{BuresNGHKKMMPSWZ13,HKPWZ15} of ASCENS is a general agile development process covering all phases of system development and relating them with the ``runtime feedback control loop'’ for awareness and adaptation. Its extension “Continuous Collaboration’’~\cite{HoelzlG16} integrates a machine-learning approach into EDLC, but as EDLC, it does not address (policy) synthesis.

The following papers address more specific development aspects.  In~\cite{Onplan15} a generic framework for modeling autonomous systems is presented which is centered around simulation-based online planning; Monte Carlo Tree Search and Cross Entropy Open Loop Planning are used for online generation of adaptive policies, but safety properties are not studied.
In \cite{DragomirIBB18}, Dragomir et al. propose an automated design process based on formal methods 
targeting partially observable timed systems. They describe how to automatically synthesize runtime monitors for fault detection, and recovery strategies for controller synthesis.

\paragraph{Multi-Objective Reinforcement Learning and Preference-Based Planning}
Many applications have not only one but several optimising goals. Multi-objective optimisation \cite{Coello18} and multi-objective reinforcement learning \cite{Drugan15} aim at simultaneously optimising several, usually conflicting objective functions. The solution is not unique, but consists of a set of so-called Pareto optimal policies which cannot be further improved in any objective without worsening at least another objective. Maintaining such multi-objective Pareto fronts is significantly more complex than dealing with a single optimizing goal. Thus in many approaches, the multi-objective problem is transformed into a single objective problem by using so-called scalarisation functions to combine the values of the different objectives into a single value. The resulting single objective problem can then be solved by standard learning and planning algorithms, for an overview see \cite{RoijersVWD13}. Using a Lagrangian approach, PsyCo transforms a multi-objective problem into a single objective problem. 

Another related approach for optimizing behavior wrt. multiple user goals (i.e. preferences) is preference-based planning and learning where (user) preferences determine the quality of plans or a policies \cite{BaierM08,SchaferH18}. Instead of using numerical rewards, policies and plans are compared w.r.t  partial order \say{preference} relations.

\paragraph{Model Checking for MDPs}
PSyCo provides a framework for synthesizing policies that maximize return while being conform to a given probabilistic requirement specification. A related line of research is treating the problems of verifying general properties of a given MDP, such as reachability. Here, verification is done either for all possible policies, or for a particular fixed one turning the MDP into a Markov chain to be verified. \cite{baier2008principles,BaierAFK18} present an overview of model checking techniques for qualitative and quantitative properties of MDPs expressed by LTL and PCTL formulas. For a recent review of this field see \citep{baier201910}. There are also software tools available, e.g. the PRISM model checker \cite{kwiatkowska2011prism}.



\paragraph{Statistical Model Checking}
Model checking is known to be time and memory consuming; its use is restricted to small and middle sized domains. Statistical model checking~\cite{legay2010statistical,LarsenL18} is a line of research for addressing this problem. 
In statistical model checking multiple executions of a system are observed and used for estimating the probabilities of system traces and giving results within confidence bounds. 
 While our work builds on these ideas, policy synthesis is not a core aspect of statistical model checking: Usually information about the verification process is not induced into a learning process \cite{legay2010statistical,clarke2011statistical,kwiatkowska2011prism}. Other prior work has discussed the Bayesian approach to model checking based on the Beta distribution, which is a key component of the PSyCo framework. In contrast to our work, these works did not use information about the verification process to guide policy synthesis \cite{jha2009bayesian,belzner2017bayesian}.

\paragraph{Safe and Robust Reinforcement Learning}
Closely related to our approach of verifiable policy synthesis are works in the area of safe reinforcement learning modeling the problem in terms of a constrained optimization problem \cite{raybenchmarking,GeZLL19,chow2018lyapunov,fan2019safety}. In contrast to our approach, these approaches do not reason about the statistical distribution of costs and corresponding constraint violation, nor do they provide a statistically grounded verification approach of given constraints.

\cite{fulton2018safe,fulton2019verifiably} propose a method for safe reinforcement learning which combines verified runtime monitoring with reinforcement learning. In contrast to our approach, their method requires a fully verified set of safe actions for a subset of the state space. While it is an interesting approach guaranteeing safety in the modeled subset, it is infeasible to perform exhaustive a-priori verification for very large or highly complex MDPs.

A notable exception is proposed in \cite{chow2017risk}, which provides statistical optimization wrt. the cost distribution tail. In contrast, PSyCo provides (a) a framework integrating formal goal specifications and policy synthesis and (b) Bayesian verification of synthesized polices including statistical confidence in verification results. While the cost distribution tail adequately captures \textit{aleatoric} uncertainty inherent to the domain, to the best of our knowledge, leveraging the Beta distribution to represent a learning agent's \textit{epistemic} confidence in constraint satisfaction and adapting the Lagrangian of a constraint optimization problem accordingly is a novel approach.

Other approaches to learning safe behavior are using a given or learned model of the environment~\cite{junges2016safety, haesaert2018temporal, hasanbeig2020cautious, Bacci2020ProbabilisticGF} or learning shields in addition to a policy assuring that only safe actions are executed~\cite{alshiekh2017safe, bharadwaj2019synthesis, avni2019run, junges2020enforcing, jansen_et_al:LIPIcs:2020:12815}. Also, there are approaches to synthesizing policies maximizing the probability of satisfying LTL constraints without maximizing reward at the same time \cite{hasanbeig2020temporal}. These approaches are orthogonal to our work, and using a learner's confidence as a learning signal in these setups could be an interesting venue for further research.

Another direction to safe reinforcement learning is the use of adversarial methods, which treat the agent's environment as an adversary to allow for synthesis of policies that are robust wrt. worst case performance or differences in simulations used for learning and real world application domains \cite{pinto2017robust, klima2019robust, phan2020learning, jaeger2020approximating}. These approaches optimize for worst-case robustness, but do not provide formal statistical guarantees on the resulting policies.

Another important line of research deals with the quantification of uncertainty and robustness to out-of-distribution data in reinforcement learning, enabling systems to identify their \say{known unknowns} \cite{sedlmeier2019uncertainty,lotjens2019safe}. This is important also from a verification perspective, as any verification results achieved before system execution are valid if the data distribution stays the same at runtime.

\section{Limitations}

%
%
%
%
%
%
%
%

While SNES is able to incorporate information from the learning process into the synthesis of feasible policies given probabilistic constraints as requirements, there are a number of limitations to be aware of.

\paragraph{Feedback loops and non-stationary data}
As the policy is changing in the course of learning, the estimation of satisfaction probability and confidence therein is done on data that is generated by a non-stationary process. In the other direction, the current estimate is used by SNES to update the policy, thus creating a feedback loop. Therefore, the estimates made by SNES while learning are to be interpreted with care: The degree of non-stationarity may severely influence the validity of the estimates. This does however not affect a posteriori verification results, which are obtained for stationary CMDP and policy.

\paragraph{Lack of convergence proof}
While our current approach leveraging the Beta distribution for adaptively adjusting the Lagrangian yields interesting and effective results empirically, SNES lacks rigorous proofs of convergence and local optimality so far. We consider this a relevant direction for future work.

\paragraph{Bounded verification}
In its current formulation, SNES performs bounded verification for a given horizon (i.e. episode length). It is unclear how to interpret or model probabilistic system requirements and satisfaction for temporally unbound systems, as in the limit every possible event will occur almost surely. A promising direction could be the integration of rates as usually performed in Markov chain analysis, or to resort to average reward formulations of reinforcement learning \cite{mahadevan1996average}. Another approach could involve learning accepting sets of B\"uchi automata based on the MDP structure with non-determinism resolved by the current policy.

\paragraph{No termination criterion}
PSyCo combines optimization goals with constraints. While it is possible to decide whether constraints are satisfied, or at least to quantify confidence in the matter, it is usually not possible to decide whether the optimization goal has been reached or not. One approach to this would be to formulate requirements wrt. reward as constraints as well, such as requiring the system to reach a certain reward threshold \cite{belzner2016qos}. In this case, policy synthesis could terminate when all given requirements are satisfied with a certain confidence.




\section{Conclusion}

We proposed to leverage epistemic uncertainty about constraint satisfaction of a reinforcement learner in safety critical domains. We introduced \textit{Policy Synthesis under probabilistic Constraints} (PSyCo), a framework for specification of requirements for reinforcement learners in constrained settings, including confidence about results. PSyCo is organized along the classical phases of systematic software development: a system specification comprising a constrained Markov decision process as domain specification and a requirement specification in terms of probabilistic constraints, an abstract design defined by an algorithm for safe policy synthesis with reinforcement learning and Bayesian model checking for system verification.

As an implementation of PSyCo we introduced \textit{Safe Neural Evolutionary Strategies} (SNES), a method for learning safe policies under probabilistic constraints. SNES is leveraging online Bayesian model checking to obtain estimates of constraint satisfaction probability and a confidence in this estimate. SNES uses the confidence estimate to weight return and cost adaptively in a principled way in order to provide a sensible optimization target wrt. the constrained task. SNES provides a way to synthesize policies that are likely to satisfy a given specification.

We have empirically evaluated SNES in a sample domain designed to show the potentially interfering optimization goals of maximizing return while reaching and maintaining constraint satisfaction. We have shown that SNES is able to synthesize policies that are very likely to satisfy probabilistic constraints.

We see various directions for future research in safe system and policy synthesis. As a direct extension to our work, it would be interesting to extend other reinforcement learning algorithms with our approach of online adaptation of the Lagrangian with Bayesian model checking, such as value-based, actor-critic and policy gradient algorithms.
We also think that notions for unbound probabilistic verification are of high interest for policy synthesis with general safety properties.
Another direction could be the inclusion of curricula into the learning process, gradually increasing the strength of the constraints over the course of learning, thus potentially speeding up the learning process and allowing for convergence to more effective local optima. 
Finally, we think that safe learning in multi-agent systems dealing with feedback loops, strategic decision making and non-stationary learning dynamics poses interesting challenges for future research.



\paragraph{Acknowledgements}
We thank the anonymous reviewers for their constructive criticisms and helpful suggestions.

\appendix

\label{sec:appendix}

\section{Finite PCTL for MDPs}\label{sub:PCTL}

We adapt probabilistic computation tree logic (PCTL) to finite sequences yielding a suitable logic to express safety constraints in our setup, allowing to specify constraints on satisfaction probabilities as well as bounding costs that may arise in system execution~\cite{pnueli1977temporal, baier2008principles}.

In our approach, a PCTL constraint is of the form $\mathbb{P}_J(\varphi)$ and specifies a bound $J \subseteq [0, 1]$ on the probability that $\varphi$ holds. The path formula $\varphi$ consists of  a single modal operator $\bigcirc$, U, $U^{\leq m}, \square$, or $\Diamond$ and propositional state formulas as arguments.
More formally, the path formula $\varphi$ is formed 
 according to the following syntax.
\begin{equation}
    \varphi = \bigcirc \phi ~\vert~ \phi_1 U \phi_2 ~\vert~ \phi_1 U^{\leq m} \phi_2 ~\vert~ \square \phi ~\vert~ \Diamond \phi
\end{equation}
Here $m \in \mathbb{N}$   and $\phi, \phi_1, \phi_2$ are propositional state formulae built over atomic formulas using the constant $\mathrm{true}$ and the propositional connectives $\neg, \wedge$, and $\vee$.

We interpret PCTL path formulas as bounded by the length of an episode and define the semantics by a satisfaction relation $\models_{\leq n}$ over finite episodes of a fixed length $n$ where w.l.o.g. we assume  $n\geq max(m, 2)$.

The semantics $\vec{e} \models_{\leq n} \varphi$ is defined as follows for finite sequences $\vec{e} = e_1, ..., e_n$ of episodes of length $n$ with $e_i = (s_{i - 1}, a_{i - 1}, s_i, r_i, c_i)$ for $i = 1, ..., n$ and $|\vec{e}| = n$. 
\begin{align}
    \vec{e} &\models_{\leq n} \bigcirc \phi
    \iff s_1 \models \phi
\\
    \vec{e} &\models_{\leq n} \phi_1 U \phi_2
    \iff \exists j,  0 \leq j \leq n  : s_j \models \phi_2 \wedge \forall k,  0 \leq k < j : s_k \models \phi_1
\\
\vec{e} &\models_{\leq n} \phi_1 U^{\leq m} \phi_2
    \iff \exists j,  0 \leq j \leq m  : s_j \models \phi_2 \wedge \forall k,  0 \leq k < j : s_k \models \phi_1
\\
    \vec{e} &\models_{\leq n} \square \phi
    \iff \forall k, 0 \leq k \leq n  : s_k \models \phi
\end{align}

The satisfaction relation $s \models \phi$ for propositional formulas $\phi$ in state $s$  is defined as usual. The semantics of the probability operator is given by the probability measure $Pr_{s}$ of all episodes of length $n$ starting at state $s$ and satisfying a path formula $\varphi$.  
\begin{equation}
  s_0 \models \mathbb{P}_{J}(\varphi) \text{  if  } Pr_{s_0}\{\vec{e} \in Episodes_n :  \vec{e} \models_{\leq n} \varphi \} \in J
\end{equation}

As usual we define  $\Diamond \phi =_\mathrm{def} \mathrm{true} ~U ~\phi$.
Then the $\square$ modality can be expressed as follows (see e.g. ~\cite{BaierAFK18}): 
$\mathbb{P}_{J}(\square \phi) =_\mathrm{def} \mathbb{P}_{[0, 1] \backslash J}(\Diamond \neg \phi)$.

This semantics coincides with the usual PCTL semantics over infinite paths. 
To show this we consider infinite sequences $\vec{e^\infty}$ of episodes and their infinite sequences $s_0, s_1, s_2, \dots$ of states 
which we denote by $s(\vec{e^\infty})$. 
Let $\vec{e^\infty} \vert_n$ denote the prefix  (or starting sequence) of length $n$ of $\vec{e^\infty}$. 
 
We define the n-bounded relativization $\varphi^{\leq n}$ of a path formula $\varphi$ (with $n \geq \max \{2, m\}$) as follows:
\begin{align}
     (\bigcirc \phi)^{\leq n} &=_{\mathrm{def}} \bigcirc \phi
\\
     (\phi_1 U \phi_2)^{\leq n} &=_{\mathrm{def}} \phi_1 U^{\leq n} \phi_2
\\
    (\phi_1 U^{\leq m} \phi_2)^{\leq n} &=_{\mathrm{def}} \phi_1 U^{\leq m} \phi_2
\\
     (\square \phi)^{\leq n} &=_{\mathrm{def}} \square^{\leq n} \phi
 \\ 
	(\Diamond \phi)^{\leq n} &=_{\mathrm{def}} \mathrm{true} ~U^{\leq n} ~\phi 
\end{align}
 
where the auxiliary box modality $\square^{\leq n}$ is semantically defined by
\begin{equation}
  s_0, s_1, s_2, \dots \models \square^{\leq n} \phi
    \iff \forall k, 0 \leq k \leq n : s_k \models \phi
\end{equation}

 For any path formula, the semantics over finite episodes of length $n$ coincides with the standard PCTL semantics of its n-bounded relativization; since we consider only bounded path formulas, the same holds for a leading probability operator: 
 
\begin{fact}
For any infinite sequence $\vec{e^\infty}$ of episodes with initial state $s_0$, any path formula $\varphi$, and  any $n$ with $n \geq \max \{2, m+1\}$, the following holds:
\begin{align}
  s(\vec{e^\infty}) \models \varphi^{\leq n}  &\iff \vec{e^\infty} \vert_n \models_{\leq n} \varphi
\\  
  s_0 \models \mathbb{P}_J(\varphi^{\leq n}) &\iff s_0 \models_{\leq n} \mathbb{P}_J(\varphi)
\end{align}
\end{fact}

\section{Cost Function and Cumulative Cost}\label{sub:cost}
For any propositional state formula $\phi$ we define a cost function   $C_\phi : S \times A \times S \rightarrow \mathbb{R}$ for a given CMDP
for all $a \in A, s, s' \in S$.
\begin{equation}
C_\phi(s, a, s') =
\begin{cases}
0               & \text{if } s' \vDash \phi  \\
1               & \text{otherwise}
\end{cases}
\end{equation}

$C_\phi$ assigns a cost for the value of $\phi$ in the post state of any transition. 

The cost of the initial state of an episode can be computed with the function $c_\phi : S \rightarrow \mathbb{R}$ defined by
\begin{equation}
c_\phi(s) =
\begin{cases}
0               & \text{if } s \vDash \phi  \\
1               & \text{otherwise}
\end{cases}
\end{equation}

Let $\vec{e}$ be an episode of length $n$, $m \leq n$, and  $\vec{s} = s_0, ..., s_{n} = s_0 \circ \vec{s'}$ denote the path $s(\vec{e})$ of that episode.

For any path $\vec{s}$, the cumulative cost function 
$\mathcal{C}_{\square \phi} : Path_n \rightarrow \mathbb{R}$ of path formula $\square \phi$ counts the number of violations of the state formula $\phi$ in $\vec{s}$; the cumulative cost is $0$ if there are no violations, if $\square \phi$ holds for $\vec{e}$. 
We give here a  specification of  $\mathcal{C}_{\square \phi}$ based on the recursively defined cost of the corresponding bounded path formula $\mathcal{C}_{\square^{\leq n} \phi}$.  
\begin{align}
\label{al:formula}
    \mathcal{C}_{\square \phi, \vec{s}} &= 
   \mathcal{C}_{\square^{\leq n} \phi, \vec{s}}
\\
    \mathcal{C}_{\square^{\leq m} \phi, \vec{s}} &=
    c_\phi(s_0) + \mathcal{C'}_{\square^{\leq m} \phi, \vec{s}}
\\
    \mathcal{C'}_{\square^{\leq m} \phi, \vec{s}} &= 
\begin{cases}
    0
    &\text{if } m = 0 \\
    C_\phi(s_0, a_0, s_1) + \mathcal{C'}_{\square^{\leq m - 1} \phi, \vec{s}} 
    &\text{if } m > 0 \\
\end{cases}
\end{align}
Clearly definition \ref{al:formula} is equivalent to the sum in equation \ref{eq:formula1}.
The cumulative of $\bigcirc \phi$ is the cost of the first action of the episode.  For $\phi_1 U \phi_2$ and $\phi_1U^{\leq m} \phi_2$ we count the number of violations of $\phi_1$ and possibly the last violation of  $\phi_2$.
\begin{align}
\mathcal{C}_{\bigcirc \phi, \vec{s}} &= C_\phi(s_0, a_0, s_1)
\\
\mathcal{C}_{\phi_1 U \phi_2, \vec{s}} &= \mathcal{C}_{ \phi_1 U^{\leq n} \phi_2, \vec{s}}
\\
\mathcal{C}_{\phi_1 U^{\leq m} \phi_2, \vec{s}} &=
    c_{\phi_2}(s_0) \left( c_{\phi_1}(s_0) + \mathcal{C'}_{\phi_1 U^{\leq m} \phi_2, \vec{s}} \right)
\\
\mathcal{C'}_{\phi_1 U^{\leq m} \phi_2, \vec{s}} &=
\begin{cases}
    1                    
    &\text{if } m = 0 \\
    C_{\phi_2}(s_0, a_0, s_1) \left( C_{\phi_1}(s_0, a_0, s_1) + \mathcal{C'}_{\phi_1 U^{\leq m - 1} \phi_2, \vec{s}} \right)
    & \text{if }  m > 0 \\
\end{cases}
\end{align}

The cumulative costs of a path constraint are $0$ if the path constraint holds for the episode. 
\begin{fact}
 Let $\vec{e}$  be any episode 
 and $\varphi$ any path formula. 
 Then the following holds:
 \begin{equation}
    \vec{e} \vDash \varphi \text{ iff } \mathcal{C}_{\varphi, s(\vec{e})} = 0
\end{equation}
\end{fact}

\end{document}